%% file: neurips_2019.tex
\documentclass{article}

\usepackage[final]{optrl_2019}

\usepackage[utf8]{inputenc} % allow utf-8 input
\usepackage[T1]{fontenc}    % use 8-bit T1 fonts
\usepackage{hyperref}       % hyperlinks
\usepackage{url}            % simple URL typesetting
\usepackage{booktabs}       % professional-quality tables
\usepackage{amsfonts}       % blackboard math symbols
\usepackage{nicefrac}       % compact symbols for 1/2, etc.
\usepackage{microtype}      % microtypography

\input{shortcuts}
\def\path2var{V^{\textrm{path}^2}}

\def\VI{\textrm{VI}}

\def\EP{\textrm{EP}}

\usepackage[normalem]{ulem}
\usepackage{xspace}

\def\regular{continuous\xspace}

\def\rregular{regular\xspace} % for (alpha,beta)-regular

\title{Continuous Online Learning and New Insights to Online Imitation Learning
} %Optimization in Episodic Markov Decision Processes}

\author{%
	Jonathan Lee\footnotemark[1] \\
	University of California, Berkeley \\
	\texttt{jonathan\_lee@berkeley.edu} \\
	\And
	Ching-An Cheng\thanks{Equal contribution} \\
	Georgia Institute of Technology\\
	\texttt{cacheng@gatech.edu} \\
	\And
	Ken Goldberg \\
	University of California, Berkeley \\
	\texttt{goldberg@berkeley.edu} \\
	\And
	Byron Boots \\
	Georgia Institute of Technology \\
	\texttt{bboots@cc.gatech.edu} \\
}

\begin{document}

\maketitle

\begin{abstract}
Online learning is a powerful tool for analyzing iterative algorithms. However, the classic adversarial setup sometimes fails to capture certain regularity in online problems in practice.
Motivated by this, we establish a new setup, called Continuous Online Learning (COL), where the gradient of online loss function changes continuously across rounds with respect to the learner's decisions. 
We show that COL covers and more appropriately describes many interesting applications, from general equilibrium problems (EPs) to optimization in episodic MDPs. 
Using this new setup, we revisit the difficulty of achieving sublinear dynamic regret. 
We prove that there is a fundamental equivalence between achieving sublinear dynamic regret in COL and solving certain EPs, and we present a reduction from dynamic regret to both static regret and convergence rate of the associated EP.
At the end, we specialize these new insights into online imitation learning and show improved understanding of its learning stability.
%optimization in episodic MDP, \red{including fitter Q-iteration and online imitation learning.}
\end{abstract}

\section{Introduction}

Online learning~\citep{gordon1999regret,zinkevich2003online} studies the interactions between a learner (i.e. an algorithm) and an opponent through regret minimization. It has proven to be a powerful framework for analyzing and designing iterative algorithms. However, while classic online learning setups focus on bounding the worst case, many applications are not naturally adversarial. 
This reality gap exists especially for iterative algorithms that are designed to solve optimization problems concerning Markov decision processes (MDPs). Because the objective is often stated in expectation for these problems, continuity properties often arise naturally from the smoothing effect of taking expectations over randomness. When such properties are ignored, theoretical analyses can be overly conservative.

To this end, we propose a new setup for online learning, called Continuous Online Learning (COL). 
In contrast to the standard adversarial setup that treats losses as adversarial, 
COL concerns online learning problems where the per-round losses change continuously with respect to the learner's decisions, and it models adversity, such as stochasticity and bias, as corruption in the feedback signals of these continuous loss sequences.
This modified setup natively captures regularity in online losses, while still being able to handle adversity that appears in common problems. As a result, certain concepts that are difficult to analyze in the classic adversarial setup (e.g. sublinear dynamic regret with an adaptive opponent) %when the opponent chooses the per-round losses adaptively based on the learner's decisions without explicit variation budgets) 
become attainable in COL.

The goal of this paper is to establish COL and to study, particularly, conditions and efficient algorithms for achieving sublinear dynamic regret. Our first result shows that achieving sublinear dynamic regret in COL, interestingly, is equivalent to solving certain equilibrium problems (EPs), which are known to be PPAD-complete\footnote{In short, they are NP problems whose solutions are known to exist, but it is open as to if they belong to P.}~\citep{daskalakis2009complexity}. In other words, achieving sublinear dynamic regret that is polynomial in the dimension of the decision set can be extremely difficult in general. Nevertheless, based on the solution concept of EP, we present a \emph{reduction} from sublinear dynamic regret to static regret and convergence to the solution of the associated EP.
This reduction allows us to quickly derive non-asymptotic dynamic regret bounds of popular online learning algorithms based on their known static regret rates.
%
%In addition, we show a reduction from general monotone EPs to COL; this result generalizes the classic reduction from convex or convex-concave problems to regret minimization~\cite{abernethy2011blackwell}.

Using these insights of COL, we revisit
online imitation learning (IL)~\citep{ross2011reduction} and show it can be framed as a COL problem.
We demonstrate that, by using standard analyses of COL, we are able to recover and improve existing understanding of online IL algorithms~\cite{ross2011reduction,cheng2018convergence,lee2018dynamic}. In particular, we characterize existence and uniqueness of solutions, and  present convergence and dynamic regret bounds for a common class of IL algorithms in deterministic and stochastic settings. A more detailed version of this paper with additional theoretical results and proofs omitted here can be found in the full technical report \citep{cheng2019online}.

\section{Continuous Online Learning}

We recall, generally, an online learning problem repeats the following steps: in round $n$, the learner plays a decision $x_n$ from a convex and compact decision set $\XX$, the opponent chooses a loss function $l_n:\XX \to \R$ based on the decisions of the learner, and then information about $l_n$ (e.g. $\nabla l_n(x_n)$) is revealed to the learner to inform the next decision. 
Classically, this abstract setup studies the \emph{adversarial} setting where $l_n$ can be almost arbitrarily chosen except for minor restrictions like convexity~\citep{shalev2012online,hazan2016introduction}. Often the performance is measured relatively through \textit{static regret},
\begin{align}  \label{eq:static regret}
\textstyle
\regret_N^s \coloneqq \sum_{n=1}^{N} l_n(x_n) - \min_{x\in\XX}\sum_{n=1}^{N} l_n(x).
\end{align}
Recently, interest has emerged in algorithms that can make nearly optimal decisions at each round. The regret is therefore measured on-the-fly and suitably named \textit{dynamic regret},
\begin{align} \label{eq:dynamic regret}
\textstyle
\regret_N^d \coloneqq \sum_{n=1}^{N} l_n(x_n) -\sum_{n=1}^{N}  l_n(x_n^*),
\end{align}
where $x_n^* \in \argmin_{x\in \XX} l_n(x)$. As dynamic regret by definition upper bounds static regret, minimizing the dynamic regret is a more difficult problem.

At a high level, one can view online learning as a protocol to describe iterative algorithms, i.e., an algorithm receives some feedback, updates its decision, tries it out and receives a performance measure, and then repeats. Indeed, this idea has made online learning a ubiquitous tool to analyze a wide range of problems. But often in these problems, the loss sequence has certain correlations; if the algorithm outputs the same decision, regardless of which iteration it is in, its performance will be measured similarly. This structure of regularity, however, is missing the classic adversarial setup. While it is possible to introduce ad-hoc constraints to limit the amount of adversity in the classic setup, as in~\cite{zinkevich2003online,mokhtari2016online,yang2016tracking,
dixit2019online,besbes2015non,jadbabaie2015online,zhang2017improved}, such a scheme often leads to case-by-case analyses and can hardly model problems where the adversity depends also on the learner's decision, like online IL of interest here (see \cref{sec:IL}). This mismatch between practice and theory makes studying certain convergence concepts difficult, such as sublinear dynamic regret which is useful to understand the performance of the last iterate produced by the algorithm.

COL differs from the classic setup mainly in the way the loss and the feedback are defined, so that it can inherently model regularity that shows up in the loss sequence of problems in practice.
 In COL, we suppose that the opponent possesses a bifunction $f:(x,x') \mapsto f_{x}(x') \in \R$, for $x,x'\in \XX$, that is \emph{unknown} to the learner.
This bifunction is used by the opponent to determine the per-round losses: 
in round $n$, if the learner chooses $x_n$, then the opponent responds with
\begin{align} \label{eq:regular per-round loss}
l_n(x) \coloneqq f_{x_n} (x).
\end{align}
Finally, the learner suffers $l_n(x_n)$ and receives feedback about $l_n$. 
For $f_{x}(x')$, we treat $x$ as the \emph{query argument}  that proposes a question (i.e. an optimization objective $f_{x}(\cdot)$), and treat $x'$ as the \emph{decision argument} whose performance is evaluated. This bifunction $f$ generally can be defined online as queried, with only one limitation that the same loss function $f_{x}(\cdot)$ must be selected by the opponent whenever the learner plays the same decision $x$. Thus, the opponent can be adaptive, but in response to only the learner's current decision. We assume, for all $x \in \XX$, $\| \nabla f_x(x) \|_* < G$ for some $G < \infty$. In~\cref{sec:IL}, we will discuss how the bifunction provides a natural interpretation for certain difficult objectives such as in online IL.

In addition to the restriction in \eqref{eq:regular per-round loss}, we impose regularity into $f$ to relate $l_n$ across rounds (so that seeking sublinear dynamic regret becomes well defined.\footnote{Otherwise the opponent can define $f_{x} (\cdot)$ pointwise for each $x$ to make $l_n(x_n) - l_n(x_n^*)$ constant.})
\begin{definition} \label{def:regular problems}
	We say an online learning problem is \emph{\regular} if $l_n$ is set as in~\eqref{eq:regular per-round loss} by a bifunction $f$  satisfying, $\forall x' \in \XX$,
	%	\begin{enumerate}\vspace{-2.5mm}
	%		\item $f_{x}(\cdot)$ is a convex function. 
	%		\item 
	$\nabla f_{x}(x')$ is a continuous map in $x$ \footnote{We define $\nabla f_{x}(x')$ as the derivative with respect to $x'$.}. 
	%	\end{enumerate}
\end{definition}
%The continuity structure in \cref{def:regular problems} and the constraint \eqref{eq:regular per-round loss} in COL limit the degree that losses can vary, making it possible for the learner to partially infer future losses from the past experiences. 
%This is similar to the predictable construction in \citep{rakhlin2013online}, which on the other hand builds on top of the classic adversarial setup. 

%%At this point, the setup of COL may sound restrictive and not of practical interest. To address this concern, we add an important \emph{nuance} in the relationship between loss and feedback in COL.
%In the classic setup, the feedback is directly determined by the loss $l_n$ and the decision $x_n$, e.g., given as full information (receiving $l_n$ or $\nabla l_n(x_n)$) or bandit (just $l_n(x_n)$). On the contrary, in COL we give the opponent the freedom to add additional stochastic or adversarial component into the feedback; e.g., in first-order feedback, the learner could receive $g_n = \nabla l_n(x_n) + \xi_n$, where $\xi_n$ is a probabilistically bounded and potentially adversarial vector, which can be used to model noise or bias in feedback. 
%%
%In other words, COL concerns loss with regularity and delegate the feedback to model adversarial and stochastic situations. This slight change in setup allows us to refine algorithm analyses, especially important in studying dynamic regret.

The continuity may appear to restrict COL to purely deterministic settings, but adversity such as stochasticity can be incorporated via an important \emph{nuance} in the relationship between loss and feedback.
In the classical online learning setting, the adversity is incorporated in the loss: the losses $l_n$ and decisions $x_n$ may themselves be generated adversarially or stochastically and then they directly determine the feedback, e.g., given as full information (receiving $l_n$ or $\nabla l_n(x_n)$) or bandit (just $l_n(x_n)$). The (expected) regret is then measured with respect to these intrinsically adversarial losses $l_n$.
By contrast, in COL, we always measure regret with respect to the true underlying bifunction $l_n = f_{x_n}$.
Instead we give the opponent the freedom to add an additional stochastic or adversarial component into the feedback; 
e.g., in first-order feedback, the learner could receive $g_n = \nabla l_n(x_n) + \xi_n$, where $\xi_n$ is a probabilistically bounded and potentially adversarial vector, which can be used to model noise or bias in feedback.
In other words, the COL setting models a true underlying loss with regularity, but allows adversity to be modeled within the feedback, analogous to stochastic feedback oracles in convex optimization. 
This additional structure is especially important for studying dynamic regret, as it allows us to always consider regret with respect to the true $f_{x_n}$ while still incorporating the possibility of stochasticity and adversity.

\section{Equivalence and Hardness of Continuous Online Learning}

We first ask what extra information the COL formulation entails. We present this result as an equivalence between achieving sublinear dynamic in COL and solving several mathematical programming problems.
%
%Before presenting bounds for dynamic regret, 
Particularly, suppose $\XX \subset \R^d$; we are interested in whether sublinear dynamic regret with polynomial dependency on $d$ 
is even possible. %for COL problems in~\cref{def:regular problems} 
It turns out, \emph{in general}, this is difficult, as least as hard as a set of difficult problems known to be PPAD-complete~\citep{daskalakis2009complexity}, even when $f_x(\cdot)$ is convex and continuous.
\begin{theorem} \label{th:hardness}
	Let $f$ be given in~\cref{def:regular problems} for a convex and compact decision set $\XX\subset\R^d$. Suppose $f_x(\cdot)$ is convex and continuous. 
	For any $f$ satisfying the above assumption, if there is an algorithm that achieves sublinear dynamic regret that is in $poly(d)$ in the associated COL, then it solves all PPAD problems in polynomial time.
	In particular, achieving sublinear dynamic regret is equivalent to solving the equilibrium problem $\EP(\XX,\Phi)$ with $\Phi(x,x') = f_x(x') - f_x(x)$ and the variational inequality $\VI(\XX, F)$ with $F(x) = \nabla f_x(x)$.
\end{theorem}

\cref{th:hardness} is an excerpt of \citep[Theorem 1]{cheng2019online} in the technical report.
We recall an EP problem, $\EP(\XX,\Phi)$, is defined by a variable set $\XX \subseteq \R^d$ and a bifunction $\Phi: \XX\times\XX\to\R$ such that\footnote{The convexity and continuity can be relaxed further, e.g., as hemi-continuity.} $\Phi(x,x)\geq0$, $\Phi(\cdot,x)$ is continuous, and $\Phi(x,\cdot)$ is convex. Its goal is to find a point $x^\* \in \XX$ such that
\begin{align*}
\Phi(x^\*,x) \geq 0, \qquad \forall x \in \XX.
\end{align*}
Similarly, the goal of a VI problem, $\VI(\XX, F)$ with $F: \XX \to \R^d$, is to find a point $x^\star \in \XX$ such that
\begin{align*}
\< F(x^\star), x - x^\star\> \geq 0, \qquad \forall x \in \XX
\end{align*} 
By definition one can see that the VI problem $\VI(\XX, F)$ is also an EP problem $\EP(\XX, \Phi_F)$ with $\Phi_F(x, x') \coloneqq \< F(x), x' - x\>$. 
%By the above definition, one can easily verify $\VI(\XX,F) = \EP(\XX, \Phi_F)$ with the bifunction $\Phi_F(x,x') = \< F(x), x' - x\>$.

In other words,~\cref{th:hardness} states that, based on the identification $\Phi(x,x') = f_x(x') - f_x(x)$ and $F(x) = \nabla f_x(x)$, achieving sublinear dynamic regret is essentially equivalent to finding an equilibrium $x^\* \in X^\*$, in which $X^\*$ denotes the set of solutions of the EP and VI (one can show these two solution sets coincide~\citep{cheng2019online}). 
Therefore, a \emph{necessary} condition for sublinear dynamic regret is that $X^\*$ is non-empty, which is true when $\nabla f_x(x)$ is continuous in $x$ and $\XX$ is compact~\cite{facchinei2007finite}.

\cref{th:hardness} also implies that extra structure on COL is necessary for designing \emph{efficient} algorithms that achieve sublinear dynamic regret and find these solutions. Specifically, we are interested in algorithms whose dynamic regret is sublinear \emph{and} polynomial in $d$. 
%\cref{th:equivalent problems} only concerns asymptotic behaviors, we learn that this is PPAD-complete; otherwise, we can use this algorithm to find an approximate solution to any Brouwer's problem in polynomial time.
The requirement of polynomial dependency is important to properly define the problem. Without it, sublinear dynamic regret can be achieved already (at least asymptotically), e.g. by simply performing a grid search that discretizes $\XX$ (as $\XX$ is compact and $\nabla f $ is continuous) albeit with an exponentially large constant.

Based on this equivalence, we can strengthen the structural properties of COL so that they are conducive to designing such efficient algorithms.
\begin{definition} \label{def:alpha-beta regular problems}
	We say a COL problem with $f$ is \emph{$(\alpha, \beta)$-{\rregular}} if for some $\alpha,\beta \in [0,\infty)$, $\forall x \in \XX$,
	\begin{enumerate}
		\item $f_{x}(\cdot)$ is a  $\alpha$-strongly convex function.
		\item $\nabla f_{\cdot}(x)$ is a $\beta$-Lipschitz continuous map.
	\end{enumerate}
	%	We say a COL problem is $(\alpha, \beta)$-\emph{\rregular} if its bifunction is .
\end{definition}
Leveraging these, we can identify similar structural properties in the equivalent problems.
\begin{proposition}\label{pr:beta-alpha strongly monotone}
	If the COL problem with $f$ is $(\alpha, \beta)$-regular, then the map $\nabla f_x(x)$ is $(\alpha - \beta)$-strongly monotone. That is, for all $x, y \in \XX$,
	\begin{align*}
		\<  \nabla f_x(x) - \nabla f_y(y), x - y\> \geq (\alpha - \beta)\| x - y\|^2
	\end{align*}
\end{proposition}
It is well known that strong monotonicity implies that $\VI(\XX, \nabla f)$ has a unique solution. 
It also implies that fast linear convergence is possible for deterministic feedback in VI problems~\citep{facchinei2007finite}.

\section{Reduction by Regularity}

We present a reduction from minimizing dynamic regret to minimizing static regret and convergence to $X^\*$. Intuitively, this is possible because Theorem~\ref{th:hardness} suggests achieving sublinear dynamic regret should not be harder than finding $x^\* \in X^\*$.  Define $\regret_N^s(x^\star) \coloneqq \sum_{n = 1}^N l_n(x_n) - l_n(x^\star)$.
%\cref{th:reduction of dynamic regret} also shows that when the dual solution set $X_\*$ of the EP problem in~\cref{th:equivalent problems} is non-empty, the dynamic regret has at least the same rate as convergence rate to $X_\*$.
\begin{theorem} \label{th:reduction of dynamic regret}
	Define $D_\XX \coloneqq \max_{x, x' \in \XX} \| x - x'\|$. Let $x^\* \in X^\*$ and $\Delta_n \coloneqq \norm{x_n - x^\*}$. If $f$ is $(\alpha, \beta)$-\rregular for $\alpha,\beta \in [0,\infty)$, then for all $N$,\vspace{-1mm}
	\begin{align*}
	\regret_N^d &\leq \min\{\textstyle G\sum_{n=1}^{N}\Delta_n, \regret_N^s(x^\star)\}  + \textstyle \sum_{n=1}^{N} \min\{\beta D_\XX\Delta_n, \frac{\beta^2}{2\alpha}\Delta_n^2\}
	\end{align*}
\end{theorem}
\cref{th:reduction of dynamic regret} roughly shows that when an equilibrium $x^\*$ exists (e.g. given by the sufficient conditions in the previous section), it provides a stabilizing effect to 
the problem, so the dynamic regret behaves almost like the static regret when the decisions are around $x^\*$.

This relationship can be used as a powerful tool for understanding the dynamic regret of existing algorithms designed for EPs and VIs. These include, e.g., mirror descent~\citep{beck2003mirror}, mirror-prox~\citep{nemirovski2004prox,juditsky2011solving}, conditional gradient descent~\citep{jaggi2013revisiting}, Mann iteration~\citep{mann1953mean}, etc. Interestingly, many of those are also standard tools in online learning with static regret bounds that are well known~\citep{hazan2016introduction}.

We can apply \cref{th:reduction of dynamic regret} in different ways, depending on the known convergence of an algorithm. For algorithms whose convergence rate of $\Delta_n$ to zero is known, \cref{th:reduction of dynamic regret} essentially shows that their dynamic regret is at most $O(  \sum_{n=1}^{N}\Delta_n)$.
For the algorithms with only known static regret bounds, we can use a corollary. %, when $\alpha > \beta > 0$. 
\begin{corollary} \label{cr:full reduction to static regret}
	If $f$ is $(\alpha,\beta)$-\rregular and $\alpha > \beta$, it holds
	$
	\regret_N^d \leq \regret_N^s(x^\star) + \frac{\beta^2 \widetilde{\regret_N^s}(x^\star) }{2\alpha(\alpha-\beta)} 
	$, 
	where {\scriptsize$\widetilde{\regret_N^s}(x^\star)$} is the static regret of the linear online learning problem with  $l_n(x) = \lr{\nabla f_n (x_n)}{x}$.
\end{corollary}
The purpose of~\cref{cr:full reduction to static regret} is not to give a tight bound, but to show that for nicer problems with $\alpha > \beta$, achieving sublinear dynamic regret is not harder than achieving sublinear static regret under linear losses. 
For tighter bounds, we still refer to~\cref{th:reduction of dynamic regret} to leverage the equilibrium convergence.
%
%We note that the results in \cref{sec:monotone ep as col} and here concern different classes of COL in general, because $\alpha > \beta$ does necessarily imply the $\EP(\XX,\Phi)$ is monotone, but only $\VI(\XX,F)$ unless $f_{x}(\cdot)$ is linear, by \cref{pr:dual solutions of EP and VI,pr:beta-alpha strongly monotone}. 

Finally, we remark \cref{th:reduction of dynamic regret} is directly applicable to expected dynamic regret (the right-hand side of the inequality will be replaced by its expectation) when the learner only has access to stochastic feedback, because the COL setup in non-anticipating. 
Similarly, high-probability bounds can be obtained based on martingale convergence theorems (see \citep{cheng2019reduction} for a COL example). In these cases, we note that the regret is defined with respect to $l_n$ in COL, \emph{not} the sampled losses.

\section{Application to Online Imitation Learning} \label{sec:IL}

In this section, we investigate an application of the COL framework in the sequential decision problem of online IL~\citep{ross2011reduction}.
We consider an episodic MDP with state space $\SS$, action space $\AA$, and finite horizon $T$. 
For any $s, s' \in \SS$ and $a \in \AA$, the transition dynamics $\PP$ gives the conditional density, denoted by $\PP(s' | s, a)$, of transitioning to $s'$ from state $s$ and action $a$. The reward of state $s$ and action $a$ is denoted as $r(s, a)$. 
A policy $\pi$ is a mapping from $\SS$ to a density over $\AA$. %, denoted by $\Delta(\AA)$. 
We suppose the MDP starts from some fixed initial state distribution. We denote the probability of being in state $s$ at time $t$ under policy $\pi$ as $d_t^\pi (s)$, and we define the average state distribution under $\pi$ as $d^\pi(s) = \frac{1}{T} \sum_{t = 1}^T d_t^\pi(s)$.
%The objectives and decision variables are problem-specific, 
%and we will discuss how they align with the COL framework in the section.

In IL, we assume that $\PP$ and $r$ are unknown to the learner, but, during training time, the learner is given access to an expert policy $\pi^\star$ and full knowledge of a supervised learning loss function $c(s, \pi; \pi^\star)$, defined for each state $s \in \SS$. 
The objective of IL is to solve
\begin{align} \label{eq:IL objective}
\min_{\pi \in \Pi} \quad \E_{s \sim d^\pi} \left[ c(s, \pi; \pi^\star) \right],
\end{align}
where $\Pi$ is the set of allowable parametric policies, which will be assumed convex; %$\subseteq \R^d$ 
note that it is often the case that $\pi^\star \not \in  \Pi$.

As $d^\pi$ is not known analytically, optimizing \eqref{eq:IL objective} directly leads to a reinforcement learning problem and therefore can be sample inefficient. \emph{Online IL}, such as the popular \textsc{DAgger} algorithm~\cite{ross2011reduction}, bypasses this difficulty by reducing \eqref{eq:IL objective} into a sequence of supervised learning problems. 
Below we describe a general construction of online IL: at the $n$th iteration
(1) execute the learner's current policy $\pi_n$ in the MDP to collect state, action samples; 
(2) update $\pi_{n+1}$ with information of the stochastic approximation of $l_n(\pi) = \E_{d^{\pi_n}}\left[ c(s, \pi; \pi^\star) \right]$ based the samples collected in the first step. Importantly, we remark that in these empirical risks, the states are sampled according to $d^{\pi_n}$ of the learner's policy.

The use of online learning to analyze online IL is well established \cite{ross2011reduction}. 
As studied in \cite{cheng2018convergence,lee2018dynamic}, these online losses can be formulated through a bifunction formulation,
$l_n(\pi) = f_{\pi_n}(\pi) = \E_{s \sim d^{\pi_n}} \left[ c(s, \pi; \pi^\star)\right]$, and the policy class $\Pi$ can be viewed as the decision set $\XX$.
Naturally, this online learning formulation results in many online IL algorithms resembling standard online learning algorithms, such as follow-the-leader (FTL), which uses full information feedback $l_n(\cdot) =\E_{s \sim  d^{\pi_n}} \left[ c(s, \cdot; \pi^\star)\right]$ at each round,  \citep{ross2011reduction} 
and mirror descent \citep{sun2017deeply}, which uses the first-order feedback $\nabla l_n(\pi_n) = \E_{d^{\pi_n}} \left[ \nabla_{\pi_n} c(s, \pi_n; \pi^\star)\right]$. This feedback can also be approximated by unbiased samples.
The original work by Ross et al.~\citep{ross2011reduction} analyzed FTL in the static regret case by immediate reductions to known static regret bounds of FTL. 
However, a crucial objective is understanding when these algorithms converge to useful solutions in terms of policy performance, which more recent work has attempted to address \cite{cheng2018convergence, lee2018dynamic, cheng2018accelerating}. According to these refined analyses, dynamic regret is a more appropriate solution concept to online IL when $\pi^\* \notin \Pi$, which is the common case in practice.

%Given an MDP with an unknown transition distribution and a supervisor policy $\pi^\star$, the objective in IL is to find a parameterized policy $\pi \in \XX \subset \R^d$ that minimizes a given supervised learning loss function $c(s, \pi ; \pi^\star)$ along rollouts generated by the learner's policy $\pi$. 
%That is, the objective is the risk $\E_{s \sim d_{\pi}} \left[ c(s, \pi; \pi^\star ) \right]$. Since $d^\pi$ is not known analytically, this problem is traditionally solved iteratively by observing risks based on the current policy's distribution $f_{\pi_n}(\pi) = \E_{s \sim d^{\pi_n}} \left[ c(s, \pi; \pi^\star ) \right]$ and using sampled feedback to update $\pi_{n+1}$.

Below we frame online IL in the proposed COL framework and study its properties based on the properties of COL that we obtained in the previous sections. 
We have already shown that the per-round loss $l_n(\cdot)$ can be written as the evaluation of a bifunction $f_{\pi_n}(\cdot)$. This COL problem is actually an $(\alpha, \beta)$-regular COL problem when the expected supervised learning loss $\E_{s\sim d^{\pi_n}}[c(s, \pi;\pi^\star)]$ is strongly convex in $\pi$ and the state distribution $d^\pi$ is Lipschitz continuous (see \cite{ross2011reduction,cheng2018convergence,lee2018dynamic}). We can then leverage our results in the COL framework to immediately answer an interesting question in the online IL problem. 
%As in \cite{ross2011reduction,cheng2018convergence,lee2018dynamic}, we strengthen the continuity and convexity assumptions with $\beta$-Lipschitz continuity of $\nabla f_\cdot(x)$ and $\alpha$-strong convexity of $f_x(\cdot)$, which corresponds to an $(\alpha, \beta)$-regular problem.
\begin{proposition}
When $\alpha > \beta$, there exists a \emph{unique} policy $\widehat \pi$ that is optimal on its own distribution:
\begin{align*}
\E_{s \sim d_{\widehat \pi_n}} \left[ c(s, \widehat \pi; \pi^\star ) \right] = 
\min_{\pi\in\Pi}  \E_{s \sim d_{\widehat \pi_n}} \left[ c(s, \pi; \pi^\star ) \right].
\end{align*}
\end{proposition}
This result is immediate from the fact that $\alpha > \beta$ implies that $\nabla f_\pi(\pi)$ is a $\mu$-strongly monotone VI with $\mu = \beta - \alpha$ by~\cref{pr:beta-alpha strongly monotone}, which is guaranteed to have a unique solution \cite{facchinei2007finite}.

Furthermore, we can improve upon the known sufficient conditions required to find this policy through online gradient descent and give a non-asymptotic convergence guarantee through a reduction to strongly monotone VIs. 
%\cheng{Is the following condition correct? Isn't this already covered by $\beta$ regularity.} 
We will additionally assume that $f$ is $L$-smooth in $\pi$, satisfying $\| \nabla f_{\pi'}(\pi_1) - \nabla f_{\pi'} (\pi_2) \| \leq L \| \pi_1 - \pi_2\|$ for any fixed query argument $\pi'$. 
%
%As in \citep{lee2018dynamic}, we assume that $d^\pi$ satisfies $\delta(d^\pi, d_{\pi'}) \leq \beta_\delta \| \pi - \pi'\|$ where $\delta: \mathbb P(s) \times \mathbb P(s) \to \R_+$ is the total variation distance. 

%This implies $f_{\pi}(\pi')$ is $(\alpha, \beta)$-regular as long as $\|\nabla c\|_\infty = \sup_{\tau, \pi} \|\nabla c(\tau, \pi; \pi^\star)\|$ is finite. 
%When $\alpha > \beta$, by \cref{th:equivalent problems} and \cref{pr:beta-alpha strongly monotone}, the VI with $F(\pi) = \E_{s \sim d_{\pi}} \left[ \nabla c(s, \pi;\pi^\star) \right]$ is $\mu$-strongly monotone with $\mu = \alpha - \beta$. 
We then apply the projection algorithm \citep{facchinei2007finite}, which is equivalent to online gradient descent studied in \citep{sun2017deeply,lee2018dynamic}. Let $P_\Pi$ denote the Euclidean projection onto $\Pi$. The online gradient descent algorithm can be described as computing the following at each round:
$
\pi_{n+1} = P_\Pi ( \pi_n - \eta_n g_n ) 
$ or equivalently
\begin{align*}
\pi_{n+1} = \argmin_{\pi \in \Pi } \ \eta_n \< g_n, \pi\> + \frac{1}{2} \| \pi - \pi_n\|^2.
\end{align*}

\begin{proposition}[\cite{facchinei2007finite}]
	If $\alpha > \beta$ and  the stepsize is chosen such that $\eta < \frac{2\mu}{(L  + \beta)^2}$, then, under the online gradient descent algorithm with deterministic feedback $g_n = \nabla l_n(\pi_n)$, it holds that
	\begin{align*}
	\| \pi_n  - \widehat \pi \|^2 \leq \left( 1 + (L + \beta)^2\eta^2 - 2\mu\eta \right)^{n - 1} \| \pi_1 - \widehat \pi\|^2
	\end{align*}
\end{proposition}

By~\cref{th:reduction of dynamic regret}, $\regret_N^d$ will therefore be sublinear (in fact $\regret_N^d = O(1)$) and the policy converges linearly to the policy that is optimal on its own distribution, $\widehat \pi$. The only condition required on the problem itself is $\alpha > \beta$ while the state-of-the-art sufficient condition of \cite{lee2018dynamic} additionally requires $\frac{\alpha}{L} > \frac{2\beta}{\alpha}$. 
The result also gives a new non-asymptotic convergence rate to $\widehat \pi$. 

The above result only considers the case when the feedback is deterministic; i.e., there is no sampling error due to executing the policy on the MDP, and the risk $\E_{d^{\pi_n}}\left[ c(s, \pi; \pi^\star) \right]$ is known exactly at each round. While this is a standard starting point in analysis of online IL algorithms \citep{ross2011reduction}, we are also interested in the more realistic stochastic case, which has so far not been analyzed for the online gradient descent algorithm in online IL. It turns out that the COL framework can be easily leveraged here too to provide a sublinear dynamic regret bound.

At round $n$, we consider observing the empirical risk $\tilde l_n (\pi) =  \frac{1}{T} \sum_{t = 1}^{T} c(s_t, \pi; \pi^\star)$ where $s_t \sim d_t^{\pi_n}$. Note that $\E [ \tilde l_n (\pi) | \pi_n ] = l_n(\pi)$ and it is easy to show that the first-order feedback $\nabla\tilde{l}_n(\pi_n)$ can be modeled as the expected gradient with an additive zero-mean noise: $g_n = \nabla l_n(\pi_n) + \epsilon_n$. For simplicity, we assume $\E\left[ \| \epsilon_n \|^2 \right] < \infty$.

\begin{proposition}
	If $\alpha > \beta$ and the stepsize is chosen as $\eta_n = \frac{1}{\sqrt n}$, then, under online gradient descent with stochastic feedback, it holds that $\E[\regret_N^d] = O(\sqrt N)$.
\end{proposition}
The proof leverages the reduction to static regret in~\cref{cr:full reduction to static regret}. It is immediate from the fact that the online IL problem is $(\alpha,\beta)$-regular (see Proposition 9 in the full technical report \cite{cheng2019online} for details). The dynamic regret is worse than that of the deterministic case, but it is still sublinear. This is the price paid for stochastically sampling from the MDP.

\section{Conclusion}

We present COL, a new class of online learning problems where the gradient varies continuously across rounds with respect to the learner's decisions. 
%This setup is motivated by the use of online learning to analyze iterative algorithms. 
We show that this setting can be equated with certain equilibrium problems (EPs) and variational inequalities (VIs). 
Leveraging this insight, we present conditions for achieving sublinear dynamic regret.
%and show a reduction from monotone EPs to COL. % with polynomial dependency on the problem's dimension. 
Furthermore, we show a reduction from dynamic regret to static regret and the convergence to equilibrium point. This insight suggests that, when these conditions are met, we may employ standard algorithms from the EP literature to achieve interpretable, sublinear dynamic regret rates.
 % in these problems. 
%
%These results reveal some core difficulties in achieving sublinear dynamic regret when there is no external constraint on the loss variation. 
Lastly, we apply our theoretical results to the online imitation learning problem, showing that interesting novel results.

%Finally we presented a generalization of the \regular problem to predictable problems that incorporate an adversarial component. 
There are several directions for future research on this topic.
Our current analyses focus on classical algorithms in online learning. We suspect that the use of adaptive or optimistic methods~\citep{cheng2018predictor} can accelerate convergence to equilibria if some coarse model of the bifunction can be estimated. This is especially relevant in applications on episodic MDPs where the expected losses are exactly determined by an underlying reward function and transition dynamics.
%We also introduce the predictable online learning setting, which generalizes the \regular problem to incorporate adversarial components.
%
In addition to online IL, there are also several iterative optimization problems with MDPs that are interesting to consider in the COL setting. First, the problems of online system identification and structured prediction have also been posed as adversarial online learning and analyzed under static regret~\citep{venkatraman2015improving,ross2011reduction}. 
We also note that the classic fitted Q-iteration \citep{gordon1995stable,riedmiller2005neural} for reinforcement learning also uses a similar setup. In round $n$, the loss can be written as 
$l_n(Q) = \E_{s,a\sim \mu_{\pi(Q_n)}} \E_{s'\sim\PP(s,a)}[(Q(s,a)- r(s,a) - \gamma \max_{a'} Q_n(s',a')   )^2]$, where $\mu_{\pi(Q_n)}$ is the state-action distribution induced by running a policy $\pi(Q_n)$ based on the Q-function estimate $Q_n$ of the learner. These problem settings can all be posed as COL problems and it would be interesting see how their algorithms and analyses can be reconciled with those of EP problems via this reduction.

%\clearpage

%\subsubsection*{Acknowledgments}
%
%Use unnumbered third level headings for the acknowledgments. All acknowledgments
%go at the end of the paper. Do not include acknowledgments in the anonymized
%submission, only in the final paper.

%\subsubsection*{References}

%References follow the acknowledgements.  Use an unnumbered third level
%heading for the references section.  Any choice of citation style is
%acceptable as long as you are consistent.  Please use the same font
%size for references as for the body of the paper---remember that
%references do not count against your page length total.

%\bibliographystyle{plain}

\bibliographystyle{unsrt}
\bibliography{neurips_2019}

\end{document}

%% file: shortcuts.tex
\usepackage{amsmath}
\usepackage{amsfonts}
\usepackage{amssymb}
\usepackage{amsthm}
\usepackage{bm}
\usepackage{bbm}
\usepackage{mathtools}
\usepackage{enumitem}
\usepackage{thmtools,thm-restate}
\usepackage{algorithm}
\usepackage{algorithmic}
\usepackage[capitalise]{cleveref}
\usepackage{color}
\usepackage[dvipsnames]{xcolor}
\usepackage{graphicx}
\usepackage{comment}
\theoremstyle{plain}
\newtheorem{theorem}{Theorem}%[section]
\newtheorem{proposition}{Proposition}%[section]
\newtheorem{corollary}{Corollary}%[section]

\theoremstyle{definition}
\newtheorem{definition}{Definition}%[section]
\theoremstyle{remark}
\def\AA{\mathcal{A}}

\def\PP{\mathcal{P}}
\def\SS{\mathcal{S}}
\def\XX{\mathcal{X}}

\def\Ebb{\mathbb{E}}

\def\Rbb{\mathbb{R}}

\def\R{\Rbb}

\def\*{\star}

\newcommand{\norm}[1]{ \| #1 \|  }

\newcommand{\lr}[2]{ \left\langle #1, #2 \right\rangle}
\newcommand{\<}{\langle}
\renewcommand{\>}{\rangle}
\DeclareMathOperator*{\argmin}{arg\,min}

\def\regret{\textrm{Regret}}

\newcommand{\E}{\Ebb}

\usepackage{etex,etoolbox}
\usepackage{environ}

\makeatletter
\providecommand{\@fourthoffour}[4]{#4}
\newcommand\fixstatement[2][\proofname\space of]{%
	\ifcsname thmt@original@#2\endcsname
	\AtEndEnvironment{#2}{%
		\xdef\pat@label{\expandafter\expandafter\expandafter
			\@fourthoffour\csname thmt@original@#2\endcsname\space\@currentlabel}%
		\xdef\pat@proofof{\@nameuse{pat@proofof@#2}}%
	}%
	\else
	\AtEndEnvironment{#2}{%
		\xdef\pat@label{\expandafter\expandafter\expandafter
			\@fourthoffour\csname #1\endcsname\space\@currentlabel}%
		\xdef\pat@proofof{\@nameuse{pat@proofof@#2}}%
	}%
	\fi
	\@namedef{pat@proofof@#2}{#1}%
}

\globtoksblk\prooftoks{1000}
\newcounter{proofcount}

\NewEnviron{proofatend}{%
	\edef\next{%
		\noexpand\begin{proof}[\pat@proofof\space\pat@label]%
			\unexpanded\expandafter{\BODY}}%
		\global\toks\numexpr\prooftoks+\value{proofcount}\relax=\expandafter{\next\end{proof}}
	\stepcounter{proofcount}}

\def\printproofs{%
	\count@=\z@
	\loop
	\the\toks\numexpr\prooftoks+\count@\relax
	\ifnum\count@<\value{proofcount}%
	\advance\count@\@ne
	\repeat}
\makeatother

\fixstatement{lemma}
\fixstatement{theorem}
\fixstatement{proposition}
\fixstatement{corollary}

%% file: neurips_2019.bbl
\begin{thebibliography}{10}

\bibitem{gordon1999regret}
Geoffrey~J Gordon.
\newblock Regret bounds for prediction problems.
\newblock In {\em Conference on Learning Theory}, volume~99, pages 29--40,
  1999.

\bibitem{zinkevich2003online}
Martin Zinkevich.
\newblock Online convex programming and generalized infinitesimal gradient
  ascent.
\newblock In {\em International Conference on Machine Learning}, pages
  928--936, 2003.

\bibitem{daskalakis2009complexity}
Constantinos Daskalakis, Paul~W Goldberg, and Christos~H Papadimitriou.
\newblock The complexity of computing a nash equilibrium.
\newblock {\em SIAM Journal on Computing}, 39(1):195--259, 2009.

\bibitem{ross2011reduction}
St{\'e}phane Ross, Geoffrey Gordon, and Drew Bagnell.
\newblock A reduction of imitation learning and structured prediction to
  no-regret online learning.
\newblock In {\em International conference on artificial intelligence and
  statistics}, pages 627--635, 2011.

\bibitem{cheng2018convergence}
Ching-An Cheng and Byron Boots.
\newblock Convergence of value aggregation for imitation learning.
\newblock In {\em International Conference on Artificial Intelligence and
  Statistics}, pages 1801--1809, 2018.

\bibitem{lee2018dynamic}
Jonathan Lee, Michael Laskey, Ajay~Kumar Tanwani, Anil Aswani, and Ken
  Goldberg.
\newblock A dynamic regret analysis and adaptive regularization algorithm for
  on-policy robot imitation learning.
\newblock In {\em Workshop on the Algorithmic Foundations of Robotics}, 2018.

\bibitem{cheng2019online}
Ching-An Cheng, Jonathan Lee, Ken Goldberg, and Byron Boots.
\newblock Online learning with continuous variations: Dynamic regret and
  reductions.
\newblock {\em arXiv preprint arXiv:1902.07286}, 2019.

\bibitem{shalev2012online}
Shai Shalev-Shwartz et~al.
\newblock Online learning and online convex optimization.
\newblock {\em Foundations and Trends{\textregistered} in Machine Learning},
  4(2):107--194, 2012.

\bibitem{hazan2016introduction}
Elad Hazan et~al.
\newblock Introduction to online convex optimization.
\newblock {\em Foundations and Trends{\textregistered} in Optimization},
  2(3-4):157--325, 2016.

\bibitem{mokhtari2016online}
Aryan Mokhtari, Shahin Shahrampour, Ali Jadbabaie, and Alejandro Ribeiro.
\newblock Online optimization in dynamic environments: Improved regret rates
  for strongly convex problems.
\newblock In {\em 2016 IEEE 55th Conference on Decision and Control (CDC)},
  pages 7195--7201. IEEE, 2016.

\bibitem{yang2016tracking}
Tianbao Yang, Lijun Zhang, Rong Jin, and Jinfeng Yi.
\newblock Tracking slowly moving clairvoyant: optimal dynamic regret of online
  learning with true and noisy gradient.
\newblock In {\em Proceedings of the 33rd International Conference on
  International Conference on Machine Learning-Volume 48}, pages 449--457.
  JMLR. org, 2016.

\bibitem{dixit2019online}
Rishabh Dixit, Amrit~Singh Bedi, Ruchi Tripathi, and Ketan Rajawat.
\newblock Online learning with inexact proximal online gradient descent
  algorithms.
\newblock {\em IEEE Transactions on Signal Processing}, 67(5):1338--1352, 2019.

\bibitem{besbes2015non}
Omar Besbes, Yonatan Gur, and Assaf Zeevi.
\newblock Non-stationary stochastic optimization.
\newblock {\em Operations research}, 63(5):1227--1244, 2015.

\bibitem{jadbabaie2015online}
Ali Jadbabaie, Alexander Rakhlin, Shahin Shahrampour, and Karthik Sridharan.
\newblock Online optimization: Competing with dynamic comparators.
\newblock In {\em Artificial Intelligence and Statistics}, pages 398--406,
  2015.

\bibitem{zhang2017improved}
Lijun Zhang, Tianbao Yang, Jinfeng Yi, Jing Rong, and Zhi-Hua Zhou.
\newblock Improved dynamic regret for non-degenerate functions.
\newblock In {\em Advances in Neural Information Processing Systems}, pages
  732--741, 2017.

\bibitem{facchinei2007finite}
Francisco Facchinei and Jong-Shi Pang.
\newblock {\em Finite-dimensional variational inequalities and complementarity
  problems}.
\newblock Springer Science \& Business Media, 2007.

\bibitem{beck2003mirror}
Amir Beck and Marc Teboulle.
\newblock Mirror descent and nonlinear projected subgradient methods for convex
  optimization.
\newblock {\em Operations Research Letters}, 31(3):167--175, 2003.

\bibitem{nemirovski2004prox}
Arkadi Nemirovski.
\newblock Prox-method with rate of convergence o (1/t) for variational
  inequalities with lipschitz continuous monotone operators and smooth
  convex-concave saddle point problems.
\newblock {\em SIAM Journal on Optimization}, 15(1):229--251, 2004.

\bibitem{juditsky2011solving}
Anatoli Juditsky, Arkadi Nemirovski, and Claire Tauvel.
\newblock Solving variational inequalities with stochastic mirror-prox
  algorithm.
\newblock {\em Stochastic Systems}, 1(1):17--58, 2011.

\bibitem{jaggi2013revisiting}
Martin Jaggi.
\newblock Revisiting frank-wolfe: Projection-free sparse convex optimization.
\newblock In {\em ICML (1)}, pages 427--435, 2013.

\bibitem{mann1953mean}
W~Robert Mann.
\newblock Mean value methods in iteration.
\newblock {\em Proceedings of the American Mathematical Society},
  4(3):506--510, 1953.

\bibitem{cheng2019reduction}
Ching-An Cheng, Remi Tachet~des Combes, Byron Boots, and Geoff Gordon.
\newblock A reduction from reinforcement learning to no-regret online learning.
\newblock {\em arXiv preprint arXiv:1911.05873}, 2019.

\bibitem{sun2017deeply}
Wen Sun, Arun Venkatraman, Geoffrey~J Gordon, Byron Boots, and J~Andrew
  Bagnell.
\newblock Deeply aggrevated: Differentiable imitation learning for sequential
  prediction.
\newblock In {\em Proceedings of the 34th International Conference on Machine
  Learning-Volume 70}, pages 3309--3318. JMLR. org, 2017.

\bibitem{cheng2018accelerating}
Ching-An Cheng, Xinyan Yan, Evangelos~A Theodorou, and Byron Boots.
\newblock Accelerating imitation learning with predictive models.
\newblock In {\em International Conference on Artificial Intelligence and
  Statistics}, 2019.

\bibitem{cheng2018predictor}
Ching-An Cheng, Xinyan Yan, Nathan Ratliff, and Byron Boots.
\newblock Predictor-corrector policy optimization.
\newblock In {\em International Conference on Machine Learning}, pages
  1151--1161, 2019.

\bibitem{venkatraman2015improving}
Arun Venkatraman, Martial Hebert, and J~Andrew Bagnell.
\newblock Improving multi-step prediction of learned time series models.
\newblock In {\em Conference on Artificial Intelligence}, 2015.

\bibitem{gordon1995stable}
Geoffrey~J Gordon.
\newblock Stable function approximation in dynamic programming.
\newblock In {\em Machine Learning Proceedings 1995}, pages 261--268. Elsevier,
  1995.

\bibitem{riedmiller2005neural}
Martin Riedmiller.
\newblock Neural fitted q iteration--first experiences with a data efficient
  neural reinforcement learning method.
\newblock In {\em European Conference on Machine Learning}, pages 317--328.
  Springer, 2005.

\end{thebibliography}
